\documentclass[runningheads]{llncs}
\usepackage[T1]{fontenc}
\usepackage{graphicx}
\usepackage[colorlinks, linkcolor=red, urlcolor=blue,
anchorcolor=red, citecolor=green]{hyperref}
\usepackage{amsfonts}
\usepackage{multirow}
\usepackage{bbding}
\usepackage{makecell}
\usepackage{authblk}
\begin{document}
\title{UniSeg: A Prompt-driven Universal Segmentation Model as well as A Strong Representation Learner
}

\titlerunning{UniSeg: A Universal Model and A Representation Learner}
%

%

\titlerunning{UniSeg: A Universal Model and A Representation Learner}
%
\author{
Yiwen Ye \inst{1 }\thanks{\small Y. Ye and Y. Xie contributed equally. Corresponding author: Y. Xia.}
\and
Yutong Xie \inst{2 }$^{\star}$
\and
Jianpeng Zhang \inst{3}
\and
Ziyang Chen \inst{1}
\and
Yong Xia\inst{1} 
}
\authorrunning{Y. Ye et al.}
\institute{National Engineering Laboratory for Integrated Aero-Space-Ground-Ocean Big Data Application Technology, School of Computer Science and Engineering, Northwestern Polytechnical University, Xi’an 710072, China \\ \and
The University of Adelaide, Australia \\ \and
DAMO Academy, Alibaba Group \\
\email{ywye@mail.nwpu.edu.cn, yutong.xie678@gmail.com, jianpeng.zhang0@gmail.com, zychen@mail.nwpu.edu.cn, yxia@nwpu.edu.cn} \\
}

\maketitle             
\begin{abstract}
The universal model emerges as a promising trend for medical image segmentation, paving up the way to build medical imaging large model (MILM).
One popular strategy to build universal models is to encode each task as a one-hot vector and generate dynamic convolutional layers at the end of the decoder to extract the interested target. 
Although successful, it ignores the correlations among tasks and meanwhile is too late to make the model `aware' of the ongoing task.
To address both issues, we propose a prompt-driven \textbf{Uni}versal \textbf{Seg}mentation model (UniSeg) for multi-task medical image segmentation using diverse modalities and domains.
We first devise a learnable universal prompt to describe the correlations among all tasks and then convert this prompt and image features into a task-specific prompt, which is fed to the decoder as a part of its input. Thus, we make the model `aware' of the ongoing task early and boost the task-specific training of the whole decoder.
Our results indicate that the proposed UniSeg outperforms other universal models and single-task models on 11 upstream tasks. 
Moreover, UniSeg also beats other pre-trained models on two downstream datasets, providing the community with a high-quality pre-trained model for 3D medical image segmentation.
Code and model are available at \href{https://github.com/yeerwen/UniSeg}{https://github.com/yeerwen/UniSeg}.

\end{abstract}
%
%

\begin{figure}[t]
\centering
\includegraphics[width=\textwidth]{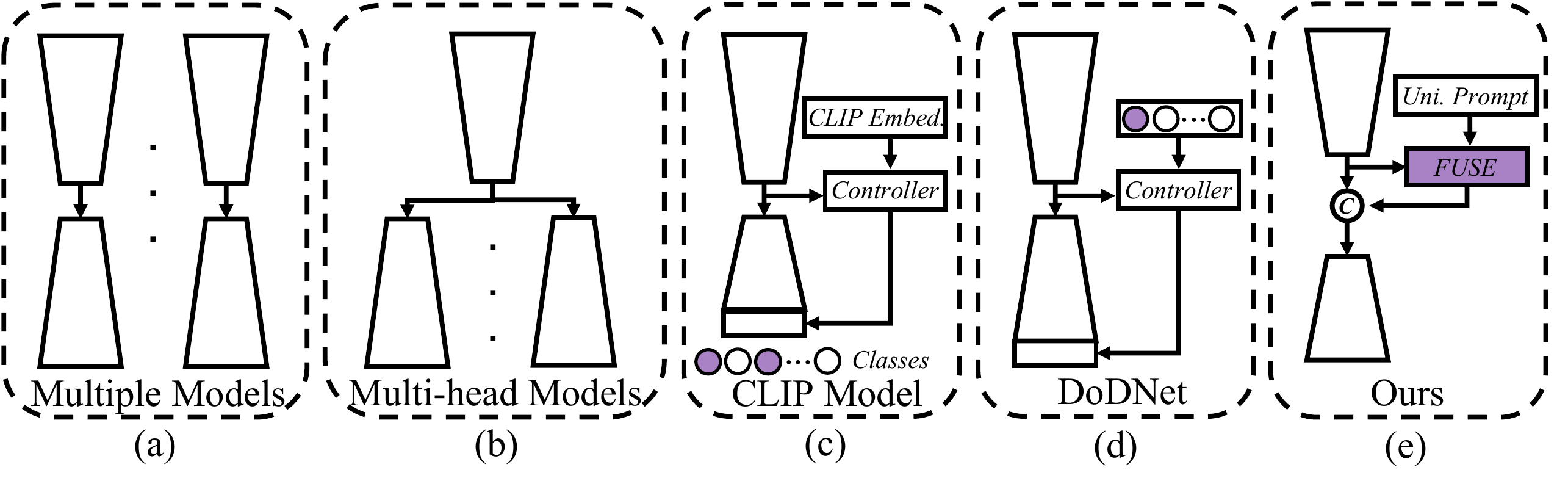}
\caption{Five strategies for multi-task medical image segmentation. (a) Multiple Models: Train $n$ models for $n$ tasks; (b) Multi-head Models: Train the model with one shared encoder and $n$ task-specific decoders; (c) CLIP-driven Model: Train one model on $n$ datasets by masking label-unavailable predictions; (d) Dynamic Convolution: Train one model on $n$ datasets using a one-hot vector as the task-related information; (e) Ours: Train one model on $n$ datasets using task-specific prompts. We use purple to highlight where to add the task-related information.}

\label{fig. vs}
\end{figure}

\section{Introduction}
Recent years have witnessed the remarkable success of deep learning in medical image segmentation. However, although the performance of deep learning models even surpasses the accuracy of human exports on some segmentation tasks, two challenges still persist. 
(1) Different segmentation tasks are usually tackled separately by specialized networks (see Fig. \ref{fig. vs}(a)), leading to distributed research efforts. (2) Most segmentation tasks face the limitation of a small labeled dataset, especially for 3D segmentation tasks, since pixel-wise 3D image annotation is labor-intensive, time-consuming, and susceptible to operator bias. 

Several strategies have been attempted to address both challenges.
First, multi-head networks (see Fig. \ref{fig. vs}(b)) were designed for multiple segmentation tasks \cite{ref6,ref3,ref30}.
A typical example is Med3D \cite{ref6}, which contains a shared encoder and multiple task-specific decoders. 
Although they benefit from the encoder parameter-sharing scheme and the rich information provided by multiple training datasets, multi-head networks are less-suitable for multi-task co-training, due to the structural redundancy caused by the requirement of preparing a separate decoder for each task.
The second strategy is the multi-class model, which formulates multiple segmentation tasks into a multi-class problem and performs it simultaneously. 
To achieve this, the CLIP-driven universal model \cite{ref7} (see Fig. \ref{fig. vs}(c)) introduces the text embedding of all labels as external knowledge, obtained by feeding medical prompts to CLIP~\cite{ref33}. 
%
However, CLIP has limited ability to generalize in medical scenarios due to the differences between natural and medical texts. It is concluded that the discriminative ability of text prompts is weak in different tasks, and it is difficult to help learn task-specific semantic information.
%
%
The third strategy is dynamic convolution. DoDNet \cite{ref1} and its variants \cite{ref2,ref4,ref5} present a universal model, which can perform different segmentation tasks based on using task encoding and a controller to generate dynamic convolutions (see Fig. \ref{fig. vs}(d)). The limitations of these models are two-fold. 
(1) Different tasks are encoded as one-hot vectors, which are mutually orthogonal, ignoring the correlations among tasks.
(2) The task-related information (\textit{i.e.}, dynamic convolution parameters) is introduced at the end of the decoder. It may be too late for the model to be `aware' of the ongoing task, making it difficult to decode complex targets.

\begin{figure}[t]
\centering
\includegraphics[width=0.85\textwidth]{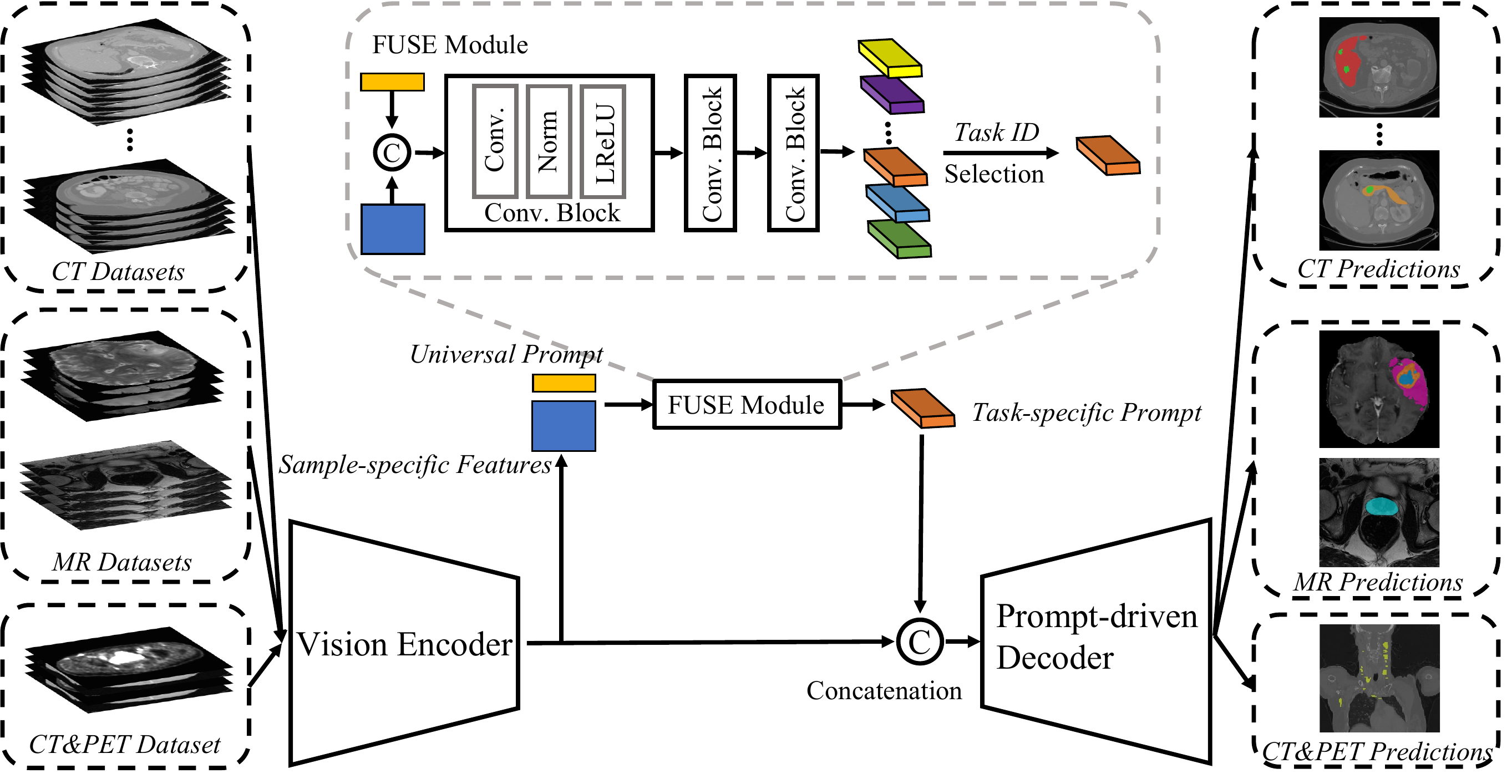}
\caption{Technical pipeline of our UniSeg, including a vision encoder, FUSE module, and a prompt-driven decoder. The sample-specific features produced by the encoder are concatenated with a learnable universal prompt as the input of the FUSE module. Then the FUSE module produces the task-specific prompt, which enables the model to be `aware' of the ongoing task.
 }

\label{fig. overview}
\end{figure}

In this paper, we propose a prompt-driven \textbf{Uni}versal \textbf{Seg}mentation model (UniSeg) to segment multiple organs, tumors, and vertebrae on 3D medical images with diverse modalities and domains.
UniSeg contains a vision encoder, a fusion and selection (FUSE) module, and a prompt-driven decoder. The FUSE module is devised to generate the task-specific prompt, which enables the model to be `aware' of the ongoing task (see Fig. \ref{fig. vs}(e)). 
Specifically, since prompt learning has a proven ability to represent both task-specific and task-invariant knowledge \cite{ref28}, a learnable universal prompt is designed to describe the correlations among tasks.
Then, the universal prompt and the features extracted by the vision encoder are fed to the FUSE module to generate task prompts for all tasks. The task-specific prompt is selected according to the ongoing task.
Moreover, to introduce the prompt information to the model early, we move the task-specific prompt from the end of the decoder to the start of the decoder (see Fig. \ref{fig. overview}). 
Thanks to both designs, we can use a single decoder and a segmentation head to predict various targets under the supervision of the corresponding ground truths. 
We collected 3237 volumetric data with three modalities (CT, MR, and PET) and various targets (eight organs, vertebrae, and tumors) from 11 datasets as the upstream dataset. 
On this dataset, we evaluated our UniSeg model against other universal models, such as DoDNet and the CLIP-driven universal model. 
We also compared UniSeg to seven advanced single-task models, such as CoTr \cite{ref8}, nnFormer \cite{ref9}, and nnUNet \cite{ref10}, which are trained independently on each dataset.
Furthermore, to verify its generalization ability on downstream tasks, we applied the trained UniSeg to two downstream datasets and compared it to other pre-trained models, such as MG \cite{ref11}, DeSD \cite{ref12}, and UniMiSS \cite{ref13}. Our results indicate that UniSeg outperforms all competing methods on 11 upstream tasks and two downstream tasks.

Our contributions are three-fold:
(1) We design a universal prompt to describe the correlations among different tasks and use it to generate task prompts for all tasks.
(2) We utilize the task-related prompt information as the input of the decoder, facilitating the training of the whole decoder, instead of just the last few layers.
(3) The proposed UniSeg can be trained on and applied to various 3D medical image tasks with diverse modalities and domains, providing a high-quality pre-trained 3D medical image segmentation model for the community.



\section{Method}
\subsection{Problem Deﬁnition}
Let $\{D_{1}, D_{2},...,D_{N}\}$ be $N$ datasets.
Here, $D_{i}=\{X_{ij},Y_{ij}\}^{n_i}_{j=1}$ represents that the $i$-th dataset has a total of $n_i$ image-label pairs, and $X_{ij}$ and $Y_{ij}$ are the image and the corresponding ground truth, respectively.
Straightforwardly, $N$ segmentation tasks can be completed by training $N$ models on $N$ datasets, respectively.
This solution, however, faces the issues of (1) designing an architecture for each task, (2) distributing research effort, and (3) dropping the benefit of rich information from other tasks.
Therefore, we propose a universal framework called UniSeg to solve multiple tasks with a single model, whose architecture was shown in Fig. \ref{fig. overview}. 
We now delve into the details of each part.

\subsection{Encoder-decoder backbone}
The main architecture of UniSeg is based on nnUNet \cite{ref10}, which consists of an encoder and a decoder shared by different tasks.
The encoder has six stages, each containing two convolutional blocks, to extract features and gradually reduce the feature resolution.
The convolutional block includes a convolutional layer followed by instance normalization and a ReakyReLU activation, and the first convolution layer of each stage is usually set to reduce the resolution with a stride of 2, except for the first stage.
To accept the multi-modality inputs, we reform the first convolution layer of the model and set up three different convolution layers to handle the input with one, two, or four channels, respectively.
After the encoder process, we obtain the sample-specific features $F \in \mathbb{R}^{C\times\frac{D}{16}\times\frac{H}{32}\times\frac{W}{32}}$, where $C$ is the number of channels and $D$, $H$, and $W$ are the depth, height, and width of the input, respectively.
Symmetrically, in each stage of the decoder, the upsampling operation implemented by a transposed convolution layer is applied to the input feature map to improve its resolution and reduce its channel number. 
The upsampled feature map is concatenated with the output of the corresponding encoder stage and then fed to a convolutional block.
After the decoder process, the output of each decoder stage is passed through a segmentation head to predict multi-scale segmentation maps for deep supervision, which is governed by the sum of the Dice loss and cross-entropy loss. Note that the channel number of multi-scale segmentation maps is set to the maximum number of classes among all tasks.

\subsection{Universal Prompt}
Following the simple idea that everything is correlated, we believe that the correlations among different segmentation tasks must exist undoubtedly, though they are ignored by DoDNet which uses a set of orthogonal and one-hot task codes.
Considering the correlations among tasks are extremely hard to handcraft, we propose a learnable prompt called universal prompt to describe them and use that prompt to generate task prompts for all tasks, aiming to encourage interaction and fusion among different task prompts.
We define the shape of the universal prompt as $F_{uni} \in \mathbb{R}^{N\times\frac{D}{16}\times\frac{H}{32}\times\frac{W}{32}}$, where $N$ is the number of tasks.

\subsection{Dynamic Task Prompt}
Before building a universal network, figuring out a way to make the model `aware' of the ongoing task is a must. 
DoDNet adopts a one-hot vector to encode each task, and the CLIP-driven universal model \cite{ref7} uses masked back-propagation to optionally optimize the task-related segmentation maps.
By contrast, we first obtain $N$ features by passing the concatenation of $F_{uni}$ and $F$ through three convolutional blocks, shown as follows
\begin{equation}\label{eq2}
\{F_{task1},F_{task2},...,F_{taskN}\}=Split(f(cat(F_{uni},F)))^{N},
\end{equation}
where $F_{taski}$ denotes the prompt features belonging to the $i$-th task, $cat(,)$ is a concatenation operation, $f(\cdot)$ denotes the feed forward process, and $Split(\cdot)^{N}$ means splitting features along the channel to obtain $N$ features with the same shape.
Then, we select the target features, called task-specific prompt $F_{tp}$, from $\{F_{task1},F_{task2},...,F_{taskN}\}$ according to the ongoing task. 
Finally, we concatenate $F$ and selected $F_{tp}$ as the decoder input.
In this way, we introduce task-related prior information into the model, aiming to boost the training of the whole decoder rather than only the last few convolution layers.

\subsection{Transfer Learning}
After training UniSeg on upstream datasets, we transfer 
the pre-trained encoder-decoder and randomly initialized segmentation heads to downstream tasks.
The model is fine-tuned in a fully supervised manner to minimize the sum of the Dice loss and cross-entropy loss.

\section{Experiments and Results}
\subsection{Datasets and Evaluation Metric}

\begin{table}[t]
  \centering
  \caption{Details of eleven upstream datasets and two downstream datasets.}
  \resizebox{1\textwidth}{!}{
    \begin{tabular}{c|cccccccc|cc|c|cc}
    \hline\hline
    \multirow{3}*{Dataset} & \multicolumn{11}{c|}{Upstream}                                & \multicolumn{2}{c}{Downstream} \\
 \cline{2-14}          & \multicolumn{8}{c|}{CT}                                       & \multicolumn{2}{c|}{MR} & CT\&PET & CT    & MR \\
 \cline{2-14}         & Liver & Kidney & HepaV & Pancreas & Colon & Lung  & Spleen & VerSe20 & Prostate & BraTS21 & AutoPET & BTCV  & VS \\
    \hline
    Organ & $\checkmark$     & $\checkmark$     & $\checkmark$     & $\checkmark$     & $\times$     & $\times$     & $\checkmark$    & $\times$     & $\checkmark$     & $\times$     & $\times$     & $\checkmark$     & $\times$ \\
    Tumor & $\checkmark$     & $\checkmark$     & $\checkmark$     & $\checkmark$     & $\checkmark$     & $\checkmark$     & $\times$     & $\times$     & $\times$     & $\checkmark$     & $\checkmark$     & $\times$     & $\checkmark$ \\
    Vertebrae & $\times$     & $\times$     & $\times$     & $\times$     & $\times$     & $\times$     & $\times$     & $\checkmark$    & $\times$     & $\times$     & $\times$     & $\times$     & $\times$ \\
    Train & 104   & 168   & 242   & 224   & 100   & 50    & 32    & 171   & 91    & 1000  & 400   & 21    & 193 \\
    Test  & 27    & 42    & 61    & 57    & 26    & 13    & 9     & 43    & 25    & 251   & 101   & 9     & 49 \\
    \hline\hline
    \end{tabular}%
    }
  \label{tab:dataset}%
\end{table}%

\begin{table}[t]
  \centering
  \caption{Results of single-task models and universal models on eleven datasets. We use Dice (\%) on each dataset and Mean Dice (\%) on all datasets as metrics. The best results on each dataset are in bold.}
    \resizebox{1\textwidth}{!}{
    \begin{tabular}{ccccccccccccc}
\hline\hline
    Method & Liver & Kidney & HepaV & Pancreas & Colon & Lung  & Spleen & VerSe20 & Prostate & BraTS21 & AutoPET & Mean \\
\hline
    \multicolumn{12}{c}{\textbf{Single-task Model}}                                               &  \\
    UNETR \cite{ref23} & 62.6  & 69.9  & 53.8  & 44.1  & 6.0   & 56.0  & 94.2  & 86.0  & 85.3  & 83.5  & 62.2  & 64.0 \\
    nnFormer \cite{ref9} & 70.7  & 80.0  & 61.3  & 57.9  & 18.8  & 66.8  & 92.2  & 84.3  & 87.0  & 82.0  & 61.0  & 69.3 \\
    PVTv2-B1 \cite{ref24}& 67.7  & 83.8  & 65.1  & 59.6  & 39.8  & 68.5  & 95.3  & 84.7  & 88.5  & 83.4  & 61.4  & 72.5 \\
    CoTr \cite{ref8}  & 74.7  & 85.1  & 67.2  & 65.8  & 33.8  & 66.9  & 95.2  & 87.1  & 88.0  & 82.9  & 58.8  & 73.2  \\
    UXNet \cite{ref26}& 75.4  & 82.2  & 67.3  & 59.4  & 39.8  & 59.5  & 95.7  & 87.1  & 88.8  & 84.3  & 68.2  & 73.4 \\
    Swin UNETR \cite{ref25}& 76.1  & 81.2  & 67.1  & 58.0  & 42.6  & 65.7  & 95.3  & 86.9  & 88.3  & 84.3  & 64.6  & 73.6 \\
    nnUNet \cite{ref10}& 77.2  & 87.5  & 69.6  & 68.8  & 49.0  & 68.4  & 96.2  & \textbf{87.2} & 89.4  & \textbf{84.4} & 64.6  & 76.6 \\
\hline
    \multicolumn{12}{c}{\textbf{Universal Model}}                                                 &  \\
    CLIP DoDNet & 62.1  & 83.6  & 57.0  & 53.3  & 19.6  & 43.8  & 51.4  & 80.2  & 89.3  & 83.1  & 65.6  & 62.6 \\
    Universal Model \cite{ref7} &74.7	&80.7	&62.2	&63.5	&52.1	&62.1	&94.5	&74.8	&87.6	&82.6	&60.0	&72.3\\
    DoDNet \cite{ref1}  & 76.7	&87.2	&70.4	&70.5	&54.6	&69.9	&\textbf{96.5}	&86.1	&89.1	&83.2	&65.3	&77.2 \\
    UniSeg  & \textbf{79.1} & \textbf{88.2} & \textbf{71.2} & \textbf{70.9} & \textbf{55.0} & \textbf{70.9}  & 96.4 & 86.1  & \textbf{89.7} & 83.3  & \textbf{69.4} & \textbf{78.2} \\
\hline\hline
    \end{tabular}%
    }
  \label{tab:upstream}%
\end{table}%

\begin{table}[t]
  \centering
  \caption{Results of self-supervised models and supervised pre-trained models. AutoPET and BraTS21 present the model per-trained on AutoPET and BraTS21 datasets, respectively. We use green numbers to indicate the performance gain of using pre-trained parameters. We repeat all experiments three times and report mean values. }
      \setlength{\tabcolsep}{3.5pt}
    \resizebox{1\textwidth}{!}{
    \begin{tabular}{ccccccccccccccc|c}
\hline\hline
    \multirow{2}*{Dataset} & \multicolumn{14}{c|}{BTCV}                                                                                     & VS \\
 \cline{2-16}          & Sp    & RKi   & LKi   & Gb    & Es    & Li    & St    & Ao    & IVC   & PSV   & Pa    & RAG   & LAG   & Mean  & Tumor \\
\hline
MG \cite{ref11} &86.8	&85.5	&83.0	&63.5	&70.5	&92.4	&78.3	&88.5	&85.3	&70.7	&71.4	&68.7	&58.2	&77.1 \textcolor[rgb]{ 0,  .69,  .314}{+2.7}	&79.3 \textcolor[rgb]{ 0,  .69,  .314}{+7.2} \\
GVSL \cite{ref31} &90.6	&92.3	&91.2	&63.7	&72.5	&95.6	&80.1	&87.5	&84.4	&71.7	&72.7	&68.1	&63.6	&79.5 \textcolor[rgb]{ 0,  .69,  .314}{+1.9}	&91.0 \textcolor[rgb]{ 0,  .69,  .314}{+2.2}\\
SMIT \cite{ref27} &90.7	&92.1	&91.9	&63.0	&74.8	&95.7	&75.9	&88.6	&86.4	&72.8	&74.3	&71.3	&69.5	&80.6 \textcolor[rgb]{ 0,  .69,  .314}{+1.3}	&92.2 \textcolor[rgb]{ 0,  .69,  .314}{+2.3} \\
UniMiSS \cite{ref13} &95.0	&92.9	&91.5	&67.1	&73.6	&96.4	&82.4	&88.9	&83.9	&73.2	&76.2	&67.1	&67.0	&81.2 \textcolor[rgb]{ 0,  .69,  .314}{+3.0}	&91.4 \textcolor[rgb]{ 0,  .69,  .314}{+2.0}\\
DeSD \cite{ref12} &96.1	&94.6	&93.2	&64.4	&75.2	&96.6	&88.7	&90.0	&87.5	&75.1	&79.9	&70.4	&70.5	&83.3 \textcolor[rgb]{ 0,  .69,  .314}{+0.8}	&92.2 \textcolor[rgb]{ 0,  .69,  .314}{+1.5} \\
\hline
AutoPET	&95.5	&93.4	&91.4	&62.8	&75.3	&96.5	&84.6	&90.0	&87.2	&75.5	&79.4	&71.2	&71.3	&82.6 \textcolor[rgb]{ 0,  .69,  .314}{-0.5}	&91.1 \textcolor[rgb]{ 0,  .69,  .314}{+0.4} \\
BraTS21	&95.9	&93.4	&90.8	&69.2	&76.5	&96.6	&84.9	&90.2	&87.6	&76.0	&80.8	&72.4	&71.7	&83.5 \textcolor[rgb]{ 0,  .69,  .314}{+0.4}	&91.2 \textcolor[rgb]{ 0,  .69,  .314}{+0.4}  \\
DoDNet \cite{ref1}	&96.4	&94.5	&89.7	&68.3	&76.9	&96.8	&86.5	&89.8	&87.7	&76.1	&81.9	&73.2	&75.2	&84.1 \textcolor[rgb]{ 0,  .69,  .314}{+0.9}	&91.8 \textcolor[rgb]{ 0,  .69,  .314}{+1.1}\\
UniSeg	&96.2	&94.4	&91.6	&68.4	&77.9	&96.7	&87.8	&90.1	&87.6	&76.7	&83.3	&73.4	&75.1	&84.6 \textcolor[rgb]{ 0,  .69,  .314}{+1.4} 	&92.9 \textcolor[rgb]{ 0,  .69,  .314}{+2.1} \\
\hline\hline

    \end{tabular}%
    }
  \label{tab:downstream}%
\end{table}%
\subsubsection{Datasets.}
For this study, we collected 11 medical image segmentation datasets as the upstream dataset to train our UniSeg and single-task models. 
The Liver and Kidney datasets are from LiTS \cite{ref14} and KiTS \cite{ref15}, respectively. 
The Hepatic Vessel (HepaV), Pancreas, Colon, Lung, and Spleen datasets are from Medical Segmentation Decathlon (MSD) \cite{ref16}.
VerSe20 \cite{ref17}, Prostate \cite{ref18}, BraTS21 \cite{ref19}, and AutoPET \cite{ref20} datasets have annotations of the vertebrae, prostate, brain tumors, and whole-body tumors, respectively.
We used the binary version of the VerSe20 dataset, where all foreground classes are regarded as one class. 
Moreover, we dropped the samples without tumors in the AutoPET dataset.

Meanwhile, We use BTCV \cite{ref21} and VS datasets \cite{ref22} as downstream datasets to verify the ability of UniSeg to generalize to other medical image segmentation tasks.
BTCV contains the annotations of 13 abdominal organs, including the spleen (Sp), right kidney (RKi), left kidney (LKi), gallbladder (Gb), esophagus (Es), liver (Li), stomach (St), aorta(Ao), inferior vena cava (IVC), portal vein and splenic vein (PSV), pancreas (Pa), right adrenal gland (RAG), and left adrenal gland (LAG).
The VS dataset contains the annotations of the vestibular schwannoma.
More details are shown in Table \ref{tab:dataset}.

\noindent{\textbf{Evaluation Metric.}}
The Dice similarity coefficient (Dice) that measures the overlap region of the segmentation prediction and ground truth is employed to evaluate the segmentation performance.

\subsection{Implementation Details}
Both pre-training on eleven upstream datasets and fine-tuning on two downstream datasets were implemented based on the nnUNet framework \cite{ref10}.
During pre-training,
we adopted the SGD optimizer and set the batch size to 2, the initial learning rate to 0.01, the default patch size to $64\times192\times192$, and the maximum training epoch to 1000 with a total of 550,000 iterations. 
In the inference stage, we employed the sliding window strategy, in which the shape of the window is the same as the training patch size, to obtain the whole average segmentation map.
During fine-tuning,
We set the batch size to 2, the initial learning rate to 0.01, the default patch size to $48\times192\times192$, and the maximum training iterations to 25,000 for all downstream datasets. 
The sliding window strategy was also employed when inference on downstream tasks.

\subsection{Results}
\noindent {\textbf{Comparing to Single-task and Universal Models.}}
Our UniSeg was compared with advanced single-task models and universal models.
The former includes UNETR \cite{ref23}, nnFormer \cite{ref9}, PVTv2-B1 \cite{ref24}, CoTr \cite{ref8}, UXNet \cite{ref26}, Swin UNETR \cite{ref25}, and nnUNet \cite{ref10}.
The latter includes DoDNet \cite{ref1}, CLIP DoDNet, which replaces the one-hot vectors with CLIP embeddings obtained by following \cite{ref7}, and CLIP-driven universal model \cite{ref7}.
For a fair comparison, the maximum training iterations of single-task models on each task are 50,000, and the patch size is $64\times192\times192$, except for Swin UNETR, whose patch size is $64\times160\times160$ due to the limitation of GPU memory.
As shown in Table \ref{tab:upstream}, Our UniSeg achieves the highest Dice on eight datasets, beating the second-best models by 1.9\%, 0.7\%, 0.8\%, 0.4\%, 0.4\%, 1.0\%, 0.3\%, 1.2\% on the Liver, Kidney, HepaV, Pancreas, Colon, Lung, Prostate, and AutoPET datasets, respectively.
Moreover, UniSeg also presents superior performance with an average margin of 1.0\% and 1.6\% on eleven datasets compared to the second-best universal model and single-task model, respectively, demonstrating its superior performance.

\noindent {\textbf{Comparing to Other Pre-trained Models.}}
We compared our UniSeg with advanced unsupervised pre-trained models, such as MG \cite{ref11}, SMIT \cite{ref27}, UniMiSS \cite{ref13}, DeSD \cite{ref12}, and GVSL\cite{ref31}, and supervised pre-trained models, such as DoDNet \cite{ref1}.
The former is officially released while the latter is trained using the datasets and backbone used in our UniSeg. 
To verify the benefit of training on multiple datasets, we also report the performance of the models per-trained on AutoPET and BraTS21, respectively.
The results in Table \ref{tab:downstream} reveal that almost all pre-trained models achieve performance gains over their baselines, which were trained from scratch.
More important, thanks to the powerful baseline and small gap between the pretext and downstream tasks, UniSeg achieves the best performance and competitive performance gains on both downstream datasets, demonstrating that it has learned a strong ability to generate image representations.
Furthermore, another advantage of UniSeg against other unsupervised pre-trained models is that it is more resource-friendly, requiring only one GPU of 11GB memory for implementation, while unsupervised pre-trained models usually require tremendous computational resources, such as eight and four V100 for UniMiSS and SMIT, respectively.

\begin{figure}[t]
\centering
\includegraphics[width=0.85\textwidth]{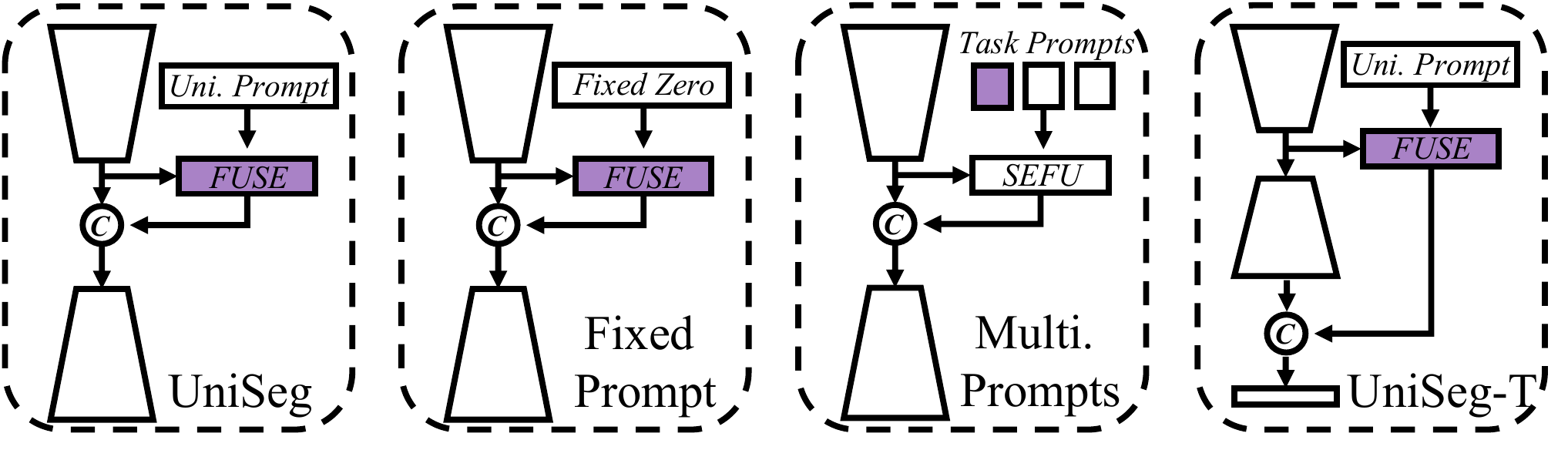}
\caption{Diagram of UniSeg, Fixed Prompt, Multiple Prompts, and UniSeg-T. Fixed Prompt initializes a zero prompt with no update. Multiple Prompts adopts multiple task-specific prompts. UniSeg-T adds the task-related prompt at the end of the decoder. We use purple to highlight where to add the task-related information.
 }
\label{fig. variants}
\end{figure}

\begin{table}[t]
  \centering
  \caption{Results of baseline, Fixed Prompt, Multiple Prompts, UniSeg-T, and our UniSeg. The baseline means the performance of our encoder-decoder backbone respectively trained on each dataset.
  We compare the mean Dice (\%) of eleven datasets.
  }
    \begin{tabular}{c|c|c|c|c|c}
\hline\hline
    Method & Baseline & Fixed Prompt & Multi. Prompts & UniSeg-T & UniSeg \\
\hline
    Dice  & 76.6  & 77.4  & 77.5  & 76.9  & 78.2 \\
\hline\hline
    \end{tabular}%
  \label{tab. variants}%
  
\end{table}%

\noindent {\textbf{Comparison of Different Variants.}}
We attempted three UniSeg variants, including Fixed Prompt, Multiple Prompts, and UniSeg-T, as shown in Fig. \ref{fig. variants}.
The results in Table \ref{tab. variants} suggest that 
(1) learnable universal prompt is helpful for building valuable prompt features;
(2) using one universal prompt instead of multiple task-independent prompts boosts the interaction and fusion among all tasks, resulting in better performance;
(3) adding task-related information in advance facilitates handling complex prediction situations.


\section{Conclusion}
This study proposes a universal model called UniSeg (a single model) to perform multiple organs, tumors, and vertebrae segmentation on images with multiple modalities and domains.
To solve two limitations existing in preview universal models, we design the universal prompt to describe correlations among all tasks and make the model `aware' of the ongoing task early, boosting the training of the whole decoder instead of just the last few layers. 
Thanks to both designs, our UniSeg achieves superior performance on 11 upstream datasets and two downstream datasets, setting a new record.

%
\bibliographystyle{splncs04}
\bibliography{UniSeg}

\begin{thebibliography}{10}
\providecommand{\url}[1]{\texttt{#1}}
\providecommand{\urlprefix}{URL }
\providecommand{\doi}[1]{https://doi.org/#1}

\bibitem{ref16}
Antonelli, M., Reinke, A., Bakas, S., Farahani, K., Landman, B.A., Litjens, G.,
  Menze, B., Ronneberger, O., Summers, R.M., van Ginneken, B., et~al.: The
  medical segmentation decathlon. arXiv preprint arXiv:2106.05735  (2021)

\bibitem{ref19}
Baid, U., Ghodasara, S., Mohan, S., Bilello, M., Calabrese, E., Colak, E.,
  Farahani, K., Kalpathy-Cramer, J., Kitamura, F.C., Pati, S., et~al.: The
  rsna-asnr-miccai brats 2021 benchmark on brain tumor segmentation and
  radiogenomic classification. arXiv preprint arXiv:2107.02314  (2021)

\bibitem{ref14}
Bilic, P., Christ, P.F., Vorontsov, E., Chlebus, G., Chen, H., Dou, Q., Fu,
  C.W., Han, X., Heng, P.A., Hesser, J., et~al.: The liver tumor segmentation
  benchmark (lits). arXiv preprint arXiv:1901.04056  (2019)

\bibitem{ref6}
Chen, S., Ma, K., Zheng, Y.: Med3d: Transfer learning for 3d medical image
  analysis. arXiv preprint arXiv:1904.00625  (2019)

\bibitem{ref33}
Conneau, A., Lample, G.: Cross-lingual language model pretraining. Advances in
  neural information processing systems  \textbf{32} (2019)

\bibitem{ref5}
Deng, R., Liu, Q., Cui, C., Asad, Z., Yang, H., Huo, Y.: Omni-seg: A single
  dynamic network for multi-label renal pathology image segmentation using
  partially labeled data. arXiv preprint arXiv:2112.12665  (2021)

\bibitem{ref3}
Fang, X., Yan, P.: Multi-organ segmentation over partially labeled datasets
  with multi-scale feature abstraction. IEEE Transactions on Medical Imaging
  \textbf{39}(11),  3619--3629 (2020)

\bibitem{ref20}
Gatidis, S., Hepp, T., Fr{\"u}h, M., La~Foug{\`e}re, C., Nikolaou, K.,
  Pfannenberg, C., Sch{\"o}lkopf, B., K{\"u}stner, T., Cyran, C., Rubin, D.: A
  whole-body fdg-pet/ct dataset with manually annotated tumor lesions.
  Scientific Data  \textbf{9}(1), ~601 (2022)

\bibitem{ref23}
Hatamizadeh, A., Tang, Y., Nath, V., Yang, D., Myronenko, A., Landman, B.,
  Roth, H.R., Xu, D.: Unetr: Transformers for 3d medical image segmentation.
  In: Proceedings of the IEEE/CVF winter conference on applications of computer
  vision. pp. 574--584 (2022)

\bibitem{ref31}
He, Y., Yang, G., Ge, R., Chen, Y., Coatrieux, J.L., Wang, B., Li, S.:
  Geometric visual similarity learning in 3d medical image self-supervised
  pre-training. In: Proceedings of the IEEE/CVF Conference on Computer Vision
  and Pattern Recognition (2023)

\bibitem{ref15}
Heller, N., Isensee, F., Maier-Hein, K.H., Hou, X., Xie, C., Li, F., Nan, Y.,
  Mu, G., Lin, Z., Han, M., et~al.: The state of the art in kidney and kidney
  tumor segmentation in contrast-enhanced ct imaging: Results of the kits19
  challenge. Medical image analysis  \textbf{67},  101821 (2021)

\bibitem{ref10}
Isensee, F., Jaeger, P.F., Kohl, S.A., Petersen, J., Maier-Hein, K.H.: nnu-net:
  a self-configuring method for deep learning-based biomedical image
  segmentation. Nature methods  \textbf{18}(2),  203--211 (2021)

\bibitem{ref27}
Jiang, J., Tyagi, N., Tringale, K., Crane, C., Veeraraghavan, H.:
  Self-supervised 3d anatomy segmentation using self-distilled masked image
  transformer (smit). In: Medical Image Computing and Computer Assisted
  Intervention. pp. 556--566. Springer (2022)

\bibitem{ref21}
Landman, B., Xu, Z., Igelsias, J., Styner, M., Langerak, T., Klein, A.: Miccai
  multi-atlas labeling beyond the cranial vault--workshop and challenge. In:
  Proc. MICCAI Multi-Atlas Labeling Beyond Cranial Vault—Workshop Challenge.
  vol.~5, p.~12 (2015)

\bibitem{ref26}
Lee, H.H., Bao, S., Huo, Y., Landman, B.A.: 3d {UX}-net: A large kernel
  volumetric convnet modernizing hierarchical transformer for medical image
  segmentation. In: The Eleventh International Conference on Learning
  Representations (2023)

\bibitem{ref7}
Liu, J., Zhang, Y., Chen, J.N., Xiao, J., Lu, Y., Landman, B.A., Yuan, Y.,
  Yuille, A., Tang, Y., Zhou, Z.: Clip-driven universal model for organ
  segmentation and tumor detection. arXiv preprint arXiv:2301.00785  (2023)

\bibitem{ref2}
Liu, P., Deng, Y., Wang, C., Hui, Y., Li, Q., Li, J., Luo, S., Sun, M., Quan,
  Q., Yang, S., et~al.: Universal segmentation of 33 anatomies. arXiv preprint
  arXiv:2203.02098  (2022)

\bibitem{ref18}
Liu, Q., Dou, Q., Yu, L., Heng, P.A.: Ms-net: multi-site network for improving
  prostate segmentation with heterogeneous mri data. IEEE transactions on
  medical imaging  \textbf{39}(9),  2713--2724 (2020)

\bibitem{ref17}
Sekuboyina, A., Husseini, M.E., Bayat, A., L{\"o}ffler, M., Liebl, H., Li, H.,
  Tetteh, G., Kuka{\v{c}}ka, J., Payer, C., {\v{S}}tern, D., et~al.: Verse: A
  vertebrae labelling and segmentation benchmark for multi-detector ct images.
  Medical image analysis  \textbf{73},  102166 (2021)

\bibitem{ref22}
Shapey, J., Kujawa, A., Dorent, R., Wang, G., Bisdas, S., Dimitriadis, A.,
  Grishchuck, D., Paddick, I., Kitchen, N., Bradford, R., et~al.: Segmentation
  of vestibular schwannoma from magnetic resonance imaging: an open annotated
  dataset and baseline algorithm. The Cancer Imaging Archive  (2021)

\bibitem{ref30}
Shi, G., Xiao, L., Chen, Y., Zhou, S.K.: Marginal loss and exclusion loss for
  partially supervised multi-organ segmentation. Medical Image Analysis
  \textbf{70},  101979 (2021)

\bibitem{ref25}
Tang, Y., Yang, D., Li, W., Roth, H.R., Landman, B., Xu, D., Nath, V.,
  Hatamizadeh, A.: Self-supervised pre-training of swin transformers for 3d
  medical image analysis. In: Proceedings of the IEEE/CVF Conference on
  Computer Vision and Pattern Recognition. pp. 20730--20740 (2022)

\bibitem{ref24}
Wang, W., Xie, E., Li, X., Fan, D.P., Song, K., Liang, D., Lu, T., Luo, P.,
  Shao, L.: Pvt v2: Improved baselines with pyramid vision transformer.
  Computational Visual Media  \textbf{8}(3),  415--424 (2022)

\bibitem{ref28}
Wang, Z., Zhang, Z., Lee, C.Y., Zhang, H., Sun, R., Ren, X., Su, G., Perot, V.,
  Dy, J., Pfister, T.: Learning to prompt for continual learning. In:
  Proceedings of the IEEE/CVF Conference on Computer Vision and Pattern
  Recognition. pp. 139--149 (2022)

\bibitem{ref4}
Wu, H., Pang, S., Sowmya, A.: Tgnet: A task-guided network architecture for
  multi-organ and tumour segmentation from partially labelled datasets. In:
  International Symposium on Biomedical Imaging. pp.~1--5. IEEE (2022)

\bibitem{ref8}
Xie, Y., Zhang, J., Shen, C., Xia, Y.: Cotr: Efficiently bridging cnn and
  transformer for 3d medical image segmentation. In: Medical Image Computing
  and Computer Assisted Intervention. pp. 171--180. Springer (2021)

\bibitem{ref13}
Xie, Y., Zhang, J., Xia, Y., Wu, Q.: Unimiss: Universal medical self-supervised
  learning via breaking dimensionality barrier. In: European Conference on
  Computer Vision. pp. 558--575. Springer (2022)

\bibitem{ref12}
Ye, Y., Zhang, J., Chen, Z., Xia, Y.: De{SD}: Self-supervised learning with
  deep self-distillation for 3d medical image segmentation. In: Medical Image
  Computing and Computer Assisted Intervention. pp. 545--555. Springer (2022)

\bibitem{ref1}
Zhang, J., Xie, Y., Xia, Y., Shen, C.: Do{D}net: Learning to segment
  multi-organ and tumors from multiple partially labeled datasets. In:
  Proceedings of the IEEE/CVF conference on computer vision and pattern
  recognition. pp. 1195--1204 (2021)

\bibitem{ref9}
Zhou, H.Y., Guo, J., Zhang, Y., Yu, L., Wang, L., Yu, Y.: nnformer: Interleaved
  transformer for volumetric segmentation. arXiv preprint arXiv:2109.03201
  (2021)

\bibitem{ref11}
Zhou, Z., Sodha, V., Pang, J., Gotway, M.B., Liang, J.: Models genesis. Medical
  image analysis  \textbf{67},  101840 (2021)

\end{thebibliography}

\newpage
\section{Appendix}

\begin{table}[htbp]
  \centering
  \caption{Implementation details of the competing models on the upstream datasets. For each model, we used the same configuration on different datasets. CE: Cross-entropy loss; BCE: Binary cross-entropy loss.}
    \resizebox{1\textwidth}{!}{
    \begin{tabular}{cccccccccc}
\hline\hline
    Method & UNETR & nnFormer & PVTv2-B1 & CoTr  & UXNet & Swin UNETR & Universal Model & DoDNet & UniSeg \\
\hline
    Initial Lr & 0.0001 & 0.01  & 0.0004 & 0.01  & 0.0001 & 0.0001 & 0.0001 & 0.01  & 0.01 \\
    Optimizer & AdamW & SGD   & AdamW & SGD   & AdamW & AdamW & AdamW & SGD   & SGD \\
    Patch Size &$64\times192^{2}$       &$64\times192^{2}$       &$64\times192^{2}$       &$64\times192^{2}$       &$64\times192^{2}$       &$64\times160^{2}$       &$64\times192^{2}$       &$64\times192^{2}$       &$64\times192^{2}$  \\
    Deep Supervision &$\checkmark$       &$\checkmark$       &$\checkmark$       &$\checkmark$       &$\checkmark$       &$\checkmark$       &$\times$       &$\times$       &$\checkmark$  \\
    Batch Size & 2     & 2     & 2     & 2     & 2     & 2     & 2     & 2     & 2 \\
    Loss  & CE+DICE & CE+DICE & CE+DICE & CE+DICE & CE+DICE & CE+DICE & BCE+DICE & CE+DICE & CE+DICE \\
    Iiteration & 50,000 & 50,000 & 50,000 & 50,000 & 50,000 & 50,000 & 50,000 & 50,000 & 50,000 \\
\hline\hline
    \end{tabular}%
    }
  \label{tab:addlabel}%
\end{table}%

\begin{table}[htbp]
  \centering
  \caption{Implementation details of the competing models on the downstream datasets. For each model, we used the same configuration on different datasets. Official: Pre-trained model from the official release.
  Implementation: Pre-trained model using our upstream datasets.}
      \resizebox{1\textwidth}{!}{
    \begin{tabular}{cccccccc}
\hline\hline
    Method & MG    & GVSL  & SMIT  & UniMiSS & DeSD  & DoDNet  & UniSeg \\
\hline
    Pre-trained Model & Official & Official & Official & Official & Official & Implementation & Implementation \\
    Initial Lr & 0.01  & 0.01  & 0.0001 & 0.0001 & 0.01  & 0.01  & 0.01 \\
    Optimizer & SGD   & SGD   & AdamW & AdamW & SGD   & SGD   & SGD \\
    Patch Size &$48\times192^{2}$         &$48\times192^{2}$         &$64\times192^{2}$       &$48\times192^{2}$       &$48\times192^{2}$       &$48\times192^{2}$       &$48\times192^{2}$  \\
    Deep Supervision &$\checkmark$       &$\checkmark$       &$\checkmark$       &$\checkmark$       &$\checkmark$       &$\checkmark$       &$\checkmark$  \\
    Batch Size & 2     & 2     & 2     & 2     & 2     & 2     & 2 \\
    Loss  & CE+DICE & CE+DICE & CE+DICE & CE+DICE & CE+DICE & CE+DICE & CE+DICE \\
    Iiteration & 25,000 & 25,000 & 25,000 & 25,000 & 25,000 & 25,000 & 25,000 \\
\hline\hline
    \end{tabular}%
    }
  \label{tab:addlabel}%
\end{table}%

\begin{table}[htbp]
  \centering
  \caption{Results of the FUSE module with 1, 2, 3, and 4 convolution blocks, respectively, on 11 upstream datasets.}
        \resizebox{1\textwidth}{!}{
    \begin{tabular}{ccccccccccccc}
\hline\hline
     Method & Liver & Kidney & HepaV & Pancreas & Colon & Lung  & Spleen & VerSe20 & Prostate & BraTS21 & AutoPET & Mean \\
\hline
    FUSE w/ 1 Block & 79.3  & 88.4  & 70.6  & 70.1  & 51.5  & 69.7  & 96.6  & 86.1  & 89.6  & 83.6 & 68.4  & 77.6 \\
    FUSE w/ 2 Blocks & 79.5 & 88.6 & 70.4  & 70.4  & 53.2  & 70.4  & 96.5 & 85.6  & 89.6  & 83.3  & 69.3  & 77.9 \\
    FUSE w/ 3 Blocks & 79.1  & 88.2  & 71.2 & 70.9 & 55.0 & 70.9 & 96.4  & 86.1 & 89.7 & 83.3  & 69.4 & 78.2 \\
    FUSE w/ 4 Blocks & 77.6  & 87.7  & 71.2  & 70.3  & 53.5  & 63.0  & 96.5  & 86.1  & 89.6  & 83.2  & 67.2  & 76.9 \\
\hline\hline
    \end{tabular}%
    }
  \label{tab:addlabel}%
\end{table}%

\begin{table}[htbp]
  \centering
  \caption{Results of the proposed universal prompt (UP) with different channel numbers, including $N$, $2\times N$, $4\times N$, and $C$, on 11 upstream datasets. $N$ is the number of upstream tasks. $C$ is the channel number of features obtained by the vision encoder. }
\resizebox{1\textwidth}{!}{
    \begin{tabular}{ccccccccccccc}
\hline\hline
     Method & Liver & Kidney & HepaV & Pancreas & Colon & Lung  & Spleen & VerSe20 & Prostate & BraTS21 & AutoPET & Mean \\
\hline
    UP w/ $N$ Channels & 79.1  & 88.2  & 71.2  & 70.9  & 55.0  & 70.9  & 96.4  & 86.1  & 89.7  & 83.3  & 69.4  & 78.2 \\
    UP w/ $2\times N$ Channels & 80.4  & 88.0  & 70.9  & 70.5  & 55.8  & 72.5  & 96.5  & 85.8  & 89.6  & 83.4  & 66.5  & 78.2 \\
    UP w/ $4\times N$ Channels & 78.0  & 87.3  & 70.9  & 70.5  & 53.7  & 69.1  & 96.4  & 85.9  & 89.7  & 83.3  & 67.7  & 77.5 \\
    UP w/ $C$ Channels & 79.0  & 88.1  & 70.7  & 70.2  & 54.3  & 67.6  & 96.5  & 85.9  & 89.6  & 83.1  & 68.8  & 77.6 \\
\hline\hline
    \end{tabular}%
}
  \label{tab:addlabel}%
\end{table}%

\begin{figure}[htbp]
\centering
\includegraphics[width=1\textwidth]{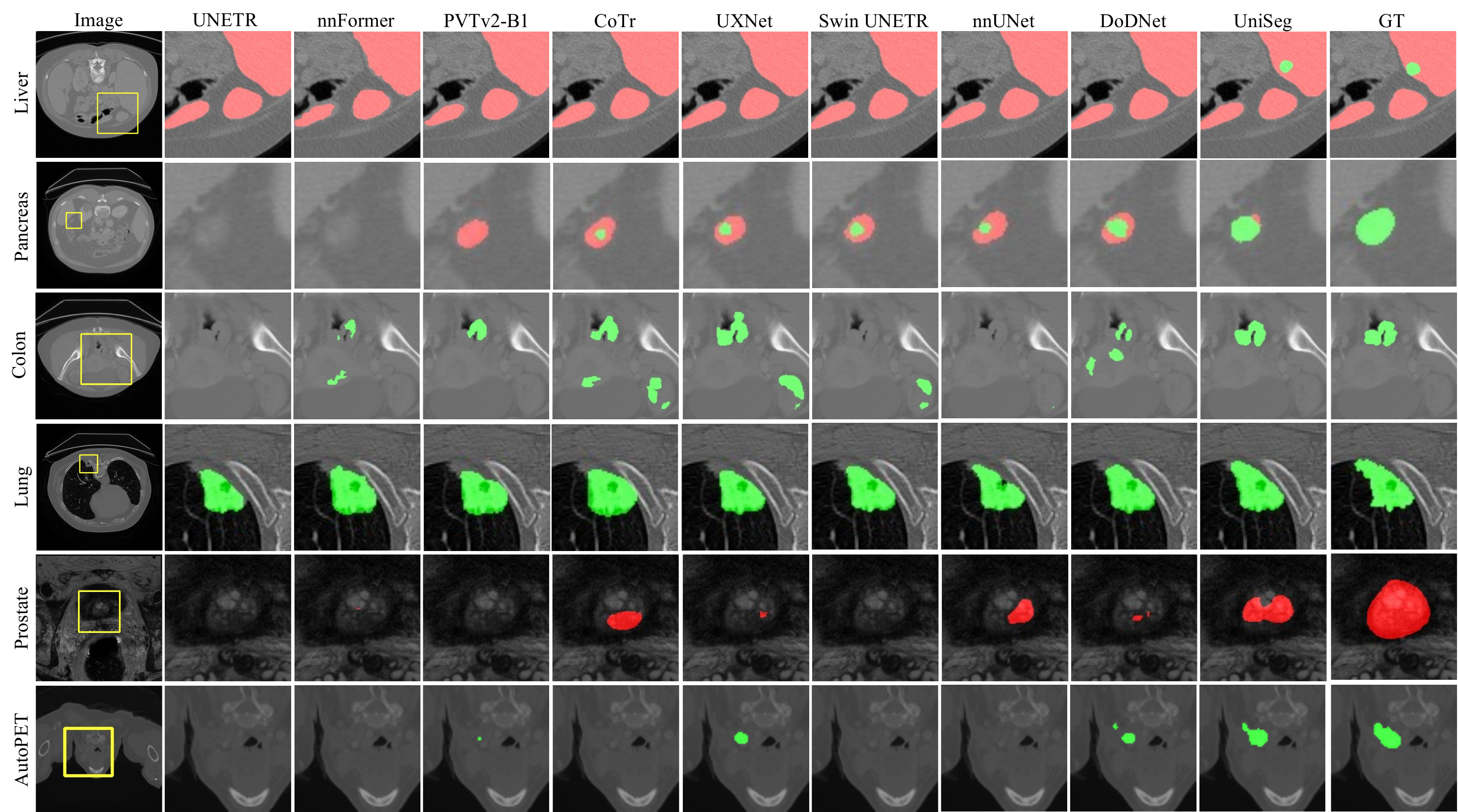}
\caption{Visualization of segmentation results obtained by UNETR, nnFormer, PVTv2-B1, CoTr, UXNet, Swin UNETR, nnUNet, DoDNet, and UniSeg on six upstream datasets, including Liver, Pancreas, Colon, Lung, Prostate, and AutoPET datasets.
 }
 \label{fig. upstream vision}
\end{figure}

\begin{figure}[htbp]
\centering
\includegraphics[width=1\textwidth]{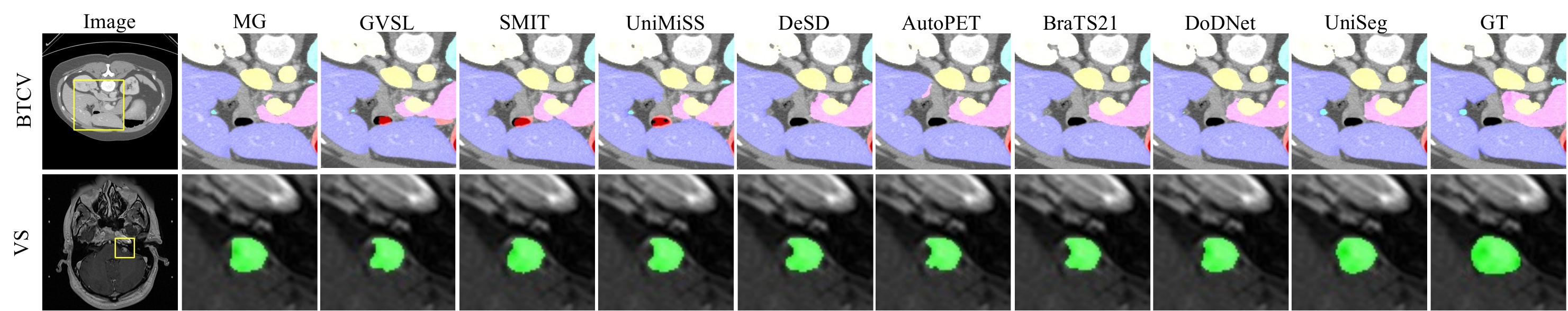}
\caption{Visualization of segmentation results obtained by MG, GVSL, SMIT, UniMiSS, DeSD, AutoPET, BraTS21, DoDNet, and UniSeg on two downstream datasets, including BTCV and VS datasets.
 }
 \label{fig. upstream vision}
\end{figure}

\end{document}